\documentclass{article}

    \PassOptionsToPackage{numbers}{natbib}
    \usepackage[final]{neurips_2023}

\usepackage[utf8]{inputenc} % allow utf-8 input
\usepackage[T1]{fontenc}    % use 8-bit T1 fonts
\usepackage{hyperref}       % hyperlinks
\usepackage{url}            % simple URL typesetting
\usepackage{booktabs}       % professional-quality tables
\usepackage{amsfonts}       % blackboard math symbols
\usepackage{nicefrac}       % compact symbols for 1/2, etc.
\usepackage{microtype}      % microtypography
\usepackage{xcolor}         % colors
\usepackage{amsmath}
\usepackage{graphicx}
\usepackage{subfig}
\usepackage{amsthm}

\title{Face-GPS: A Comprehensive Technique for Quantifying Facial Muscle Dynamics in Videos}

\author{%
  Juni Kim\thanks{Equally contributed.} \thanks{Corresponding author.} \\
  Stanford Online High School\\
  Redwood City, CA, 94063 \\
  \texttt{unickim@ohs.stanford.edu} \\
  \And
  Zhikang Dong\footnotemark[1] \\
  Department of Applied Mathematics and Statistics \\
  Stony Brook, NY, 11794 \\
  \texttt{zhikang.dong.1@stonybrook.edu} \\
  \And
  Pawe\l \ Polak \\
  Department of Applied Mathematics and Statistics \\
  Institute for Advanced Computational Science\\
  Stony Brook, NY, 11794 \\
  \texttt{pawel.polak@stonybrook.edu} \\
}

\begin{document}

\maketitle

\begin{abstract}
   We introduce a novel method that combines differential geometry, kernels smoothing, and spectral analysis to quantify facial muscle activity from widely accessible video recordings, such as those captured on personal smartphones. Our approach emphasizes practicality and accessibility. It has significant potential for applications in national security and plastic surgery. Additionally, it offers remote diagnosis and monitoring for medical conditions such as stroke, Bell's palsy, and acoustic neuroma. Moreover, it is adept at detecting and classifying emotions, from the overt to the subtle. The proposed face muscle analysis technique is an explainable alternative to deep learning methods and a non-invasive substitute to facial electromyography (fEMG).
\end{abstract}

\section{Introduction}

The increasing adoption of facial recognition technology by companies, governments, and consumers for various purposes—including marketing, surveillance, security, identification, and personal convenience—has amplified the importance of precise and efficient face analysis. As a fundamental aspect of human communication, facial expressions convey emotions, feelings, and personal identities. Traditional facial muscle movement or expression analysis mainly employs facial electromyography (fEMG) to detect emotion-related muscle contractions and relaxations~\citep{ekman2003unmasking,sato2020physiological,yadav2019emotional,xi2020facial,kehri2019analysis}. However, fEMG's need for specialized equipment and expertise renders it inflexible and unsuitable for quick prediagnosis. As an alternative, the Facial Action Coding System (FACS) uses visualization-based methods to categorize facial actions into Action Units (AUs)\citep{ekman1978facial,ekman2002facial}. Despite its capability to capture distinct facial expressions through unique muscle combinations\citep{ekman1978facial, gosselin1995components, kohler2004differences}, FACS is time-consuming, subject to bias, and unsuitable for large sample studies. To mitigate these limitations, researchers have explored automated scoring systems employing techniques like multi-resolution Haar wavelet basis and hierarchical AdaBoost cascade classifier~\citep{viola2004robust}, probabilistic likelihood classifiers~\citep{wang2008automated}, and Dynamic Bayesian Networks for AU modeling~\citep{li2013unified}.

Recently, numerous deep learning approaches have been proposed for video-based face detection and facial action analysis tasks.~\citep{le2015simple} utilized an identity matrix to initialize RNNs, addressing the exploding and vanishing gradient problems~\citep{ebrahimi2015recurrent}.~\citep{yu2018spatio} implemented a nested LSTM, composed of two sub-LSTMs, T-LSTM and C-LSTM, with the former modeling the temporal dynamics of spatio-temporal features and the latter combining the output of the T-LSTM to extract multi-representations. Convolutional neural networks (CNNs)~\citep{lecun1998gradient} and their extension, 3D CNNs, have also been extensively employed in such tasks~\citep{ouyang2017audio, abbasnejad2017using, fan2016video, al2018deep, barros2016developing}. 3D CNNs can also serve as feature extractors~\citep{pini2017modeling} for multimodal learning. While deep-learning-based methods show promising results, their lack of explainability poses challenges for those without domain expertise. Although research has sought to interpret facial recognition outcomes~\citep{saadon2023real,williford2020explainable,franco2022deep,fu2022explainability,rajpal2023xai}, these studies have neither considered the importance of facial muscle movements for clinical applications nor effectively eliminated the confounding effects of background noise and head movements.

Initially developed for non-contact assessments of material mechanical properties and the detection of micro-movements on material surfaces, Digital Image Speckle Correlation (DISC) has recently found medical applications, owing to the traceable patterns of pores on human skin. DISC has been utilized to measure skin sample deformation \cite{Guanetal:04}, compare dermal substitutes \cite{fritz2012comparison}, provide diagnostic and prognostic data for managing and treating vestibular schwannomas (acoustic neuroma) \cite{BhatnagarFioreRafailovichDavis:14}, and identify optimal Botox injection sites \cite{Bhatnagaretal:13, Vermaetal:19}. In static environments, DISC has been employed to accurately analyze facial muscle movements and classify corresponding facial expressions using a short series of 2D images when the patient's head remains stationary across frames \cite{saadon2023real, Wangetal:14, Pamudurthyetap:05}. A related technique, optical flow, has been widely incorporated in computer image analysis to track moving objects within videos and deep neural networks. Inspired by \cite{simonyan2014two}, \cite{sun2019deep} proposed a multi-input network that extracted spatial information from face images and temporal information from optical flow between emotional and neutral faces, investigating three distinct feature fusion strategies. This approach leveraged revised optical flow information to measure muscle changes and employed a deep multi-task learning network to detect micro-expressions. Although facial landmark trajectories can accurately measure facial muscle changes, they are sparse and primarily limited to detecting specific facial parts (e.g., eyes, nose, and mouth), which may lead to a loss of information.

In this paper, we present a new algorithm for quantifying facial muscle movements using standard videos. Initially, we extract face manifolds from video frames and convert them into a canonical face representation to minimize the effects of background and head movements. Next, we apply a smoothing mechanism to improve the DISC results. To enhance the interpretability of facial muscle movements, these refined DISC results are further smoothed using multiple kernels and are then added to the original videos for expert identification and diagnosis.
\section{Methodology}
\label{sec:Methodology}
We represent a grayscale video as a sequence of $p$ frames, denoted by $\mathbf{V} = \{V_i\}_{i=1}^p$, where each frame $V_i \in \mathbb{R}^{\mathcal{N} \times \mathcal{M}}$ is a matrix corresponding to the video's resolution, specifically $\mathcal{N} \times \mathcal{M}$. For every frame, denoted as $V_i$, we introduce $F_i$ as a smooth, path-connected manifold of $n$ dimensions. Each manifold $F_i$ can be expressed as $F_i = \{(X_j^{(i)}, E_j^{(i)}) \}_{j=1}^\ell$, which constructs a graph comprising $\ell$ unique landmarks, denoted as $X_j^{(i)}$, and edges referred to as $E_j^{(i)}$. This graphical representation, derived from connecting a grid of landmarks, incorporates $K$ triangles into each individual frame. The manifold $F_i$ is assumed to be intrinsic, representing all potential latent states of a lower-dimensional system. We use Mediapipe~\citep{grishchenko2020attention} to extract canonical faces involving a new set of landmarks, $\widetilde{X}_j^{(i)} \in \mathbb{R}^{n}$. Since $F_i$ is a smooth $n$-manifold, it is possible to identify a smooth coordinate chart $(U_i, \varphi_k)$ within a specific triangle $\Delta_k$ situated in the manifold, where $U_i$ is an open subset of $F_i$ and $\varphi_k: U_i \rightarrow \widetilde{U}_i$ is a homeomorphism
from $U_i$ to an open subset $\widetilde{U}_i=\varphi_k(U_i) \subseteq \mathbb{R}^2$. 
Consider the coordinates of the vertices $X_j^{(i)} \subseteq U_i$ in the video frames, and their corresponding vertices $\widetilde{X}^{(i)}_j \subseteq \widetilde{U}_i$ within the canonical face model. We can find the local affine transformation $\varphi_k$ to project all the pixels inside the given triangle  $\Delta_k$ in the video frames to the corresponding triangle $\widetilde{\Delta}_k$ in the canonical frames. We obtain a sequence of canonical frames from the video $\widetilde{\mathbf{F}}=\{ \widetilde{F}_i\}_{i=1}^p$. These canonical embeddings allow us to measure facial muscle movements, even when the face is in motion or turning to a side.

We then measure face movement on the canonical face using DISC on a pair of corresponding canonical embeddings of face manifolds from two frames from the video via Lucas-Kanade algorithm ~\citep{lucas1981iterative} sparse optical flow in consecutive frames. In the canonical face, $\forall (\widetilde{x}_j, \widetilde{y}_j) \in \widetilde{F}_j$ have the corresponding optical flow $(\widetilde{x}_j + d_{\widetilde{x}_j}, \widetilde{y}_j + d_{\widetilde{y}_j}) \in \widetilde{F}_j$. Facial muscle movements are then measured by analyzing the pixel displacements within the corresponding facial regions. We convert $(d_{\widetilde{x}_j}, d_{\widetilde{y}_j})$ into polar coordinate system $(\widetilde{\mathbf{r}}_j, \widetilde{\mathbf{\theta}}_j)$. We then perform spectral analysis on $\widetilde{\mathbf{r}}_j$ to smooth length based on geometric features, and we use a wavelet smoothing algorithm on the angle changes.

Inspired by FACS~\citep{ekman1978facial}, we introduce a Multiple Kernel Smoothing (MKS) approach that combines Gaussian RBF Kernels from specific facial muscle descriptors for feature-selective noise reduction and amplification of true facial muscle movements within the face manifold \( \mathbf{F} \). 

Suppose we have $m$ different facial muscles descriptor $\{D_1,D_2,\ldots,D_m\}$ in frame $V_i$. Based on Theorem 6.1 and Theorem 6.2 in ~\citep{jayasumana2015kernel}, we never need to calculate geodesic distance on the face manifold, the Euclidean distance gives the equivalent results since the algorithm only requires the inner product of the Gaussian RBF Kernels. We then have Gaussian RBF kernel on Euclidean space $k_f^{(i)}:=\exp \left(-\gamma d_e^2 (\widetilde{X}^{(i)}_j, D_i) \right)$, for $j=1,2,\ldots,l$, where $d_e$ is the Euclidean distance. 

Thus our MKS approach is given as 

\[\widetilde{\mathbf{r}}^{\prime\prime}_j = \frac{1}{m} \sum_{i=1}^m w_i\widetilde{\mathbf{r}}^\prime_jk_f^{(i)},
\]
where \( w_i \) represents the weight assigned to each facial muscle descriptor, computed using any deep-learning models for face expression recognition.

We revert \((\widetilde{\mathbf{r}}^{\prime\prime}_j, \widetilde{\mathbf{\theta}}^\prime_j)\) back to the Cartesian coordinate system \((d_{\widetilde{x}^\prime_j}, d_{\widetilde{y}^\prime_j})\) and perform the inverse affine transformation to obtain the coordinates of the smoothed optical flow. This process enables us to quantify the displacement of facial muscles in a given video.
\section{Empirical Results}
To evaluate the effectiveness of our method, we generate a dynamic vector field similar to a heatmap, which is superimposed onto the participant videos. This visual representation clearly indicates both the magnitude and direction of facial muscle movements during the recording. Using 468 facial landmarks and 854 manifold triangles from Mediapipe~\citep{grishchenko2020attention}, we want to improve accuracy by sampling more landmarks in each triangle. Finally, we have 3,681 landmarks that cover all facial muscles.

\begin{figure}[t]
\centering
{%
  \includegraphics[height=25mm, width=.25\linewidth]{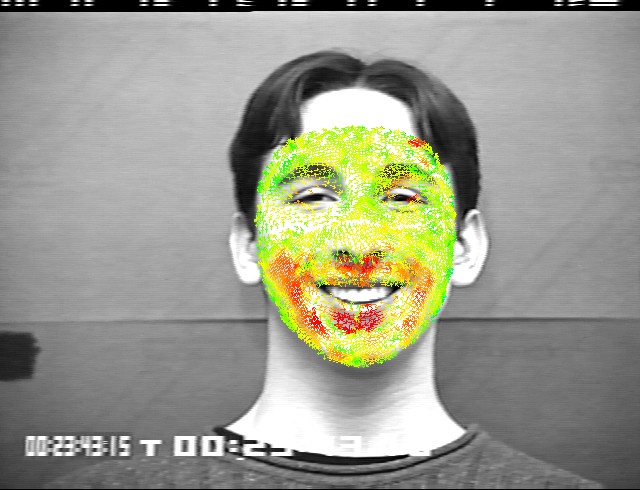}%
}\hfill
{%
  \includegraphics[height=25mm, width=.25\linewidth]{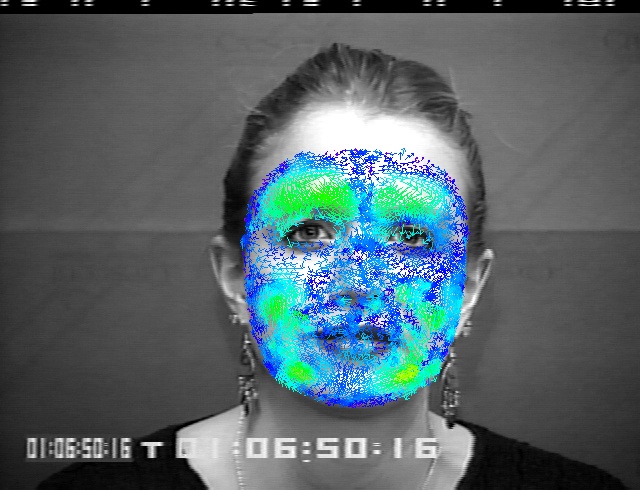}%
}\hfill
% {%
%   \includegraphics[height=20mm, width=.2\linewidth]{figures/happy/imap001313_eigen.jpg}%
% }\hfill
{%
  \includegraphics[height=25mm, width=.25\linewidth]{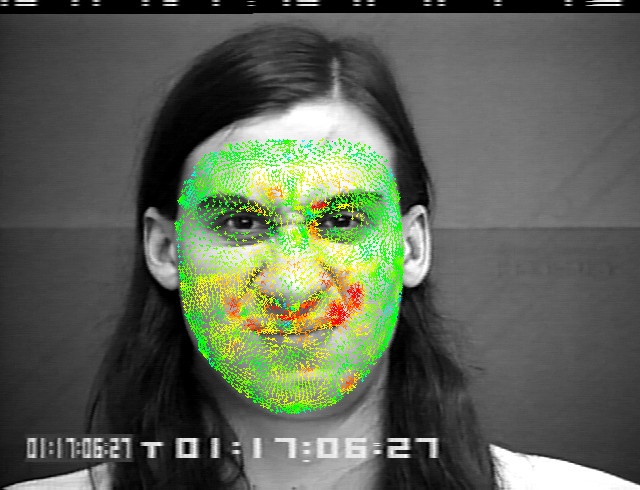}%
}\hfill
{%
  \includegraphics[height=25mm,width=.25\linewidth]{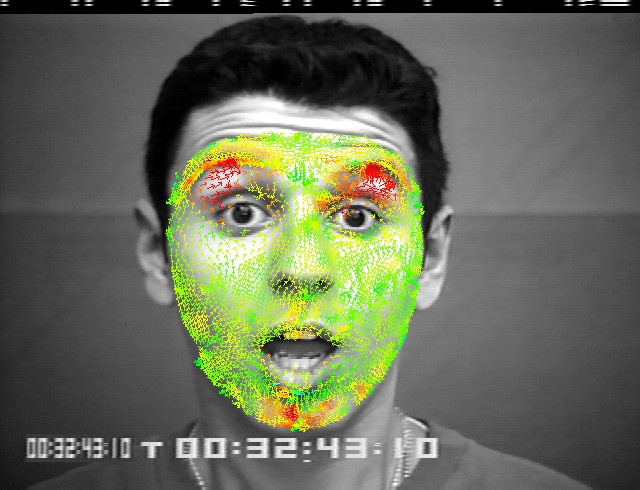}%
}
{%
  \includegraphics[height=25mm, width=.25\linewidth]{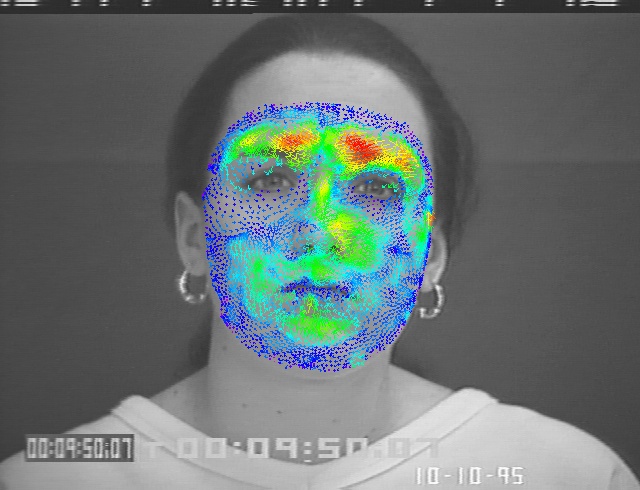}%
}\hfill
{%
  \includegraphics[height=25mm, width=.25\linewidth]{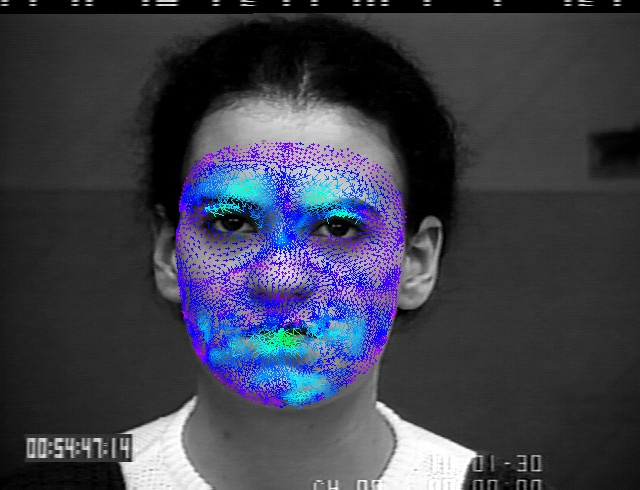}%
}\hfill
{%
  \includegraphics[height=25mm, width=.25\linewidth]{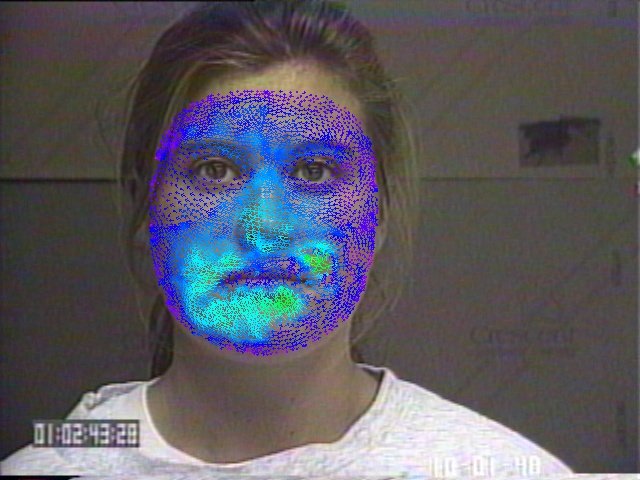}%
}\hfill
\caption{Our method applied on videos exhibiting Happy, Fear, Disgust, Surprise, Sad, Anger, and Contempt from left to right.}
\label{fig:emotions}
\end{figure}

Our study utilizes the CK+ Dataset \citep{ckplus}, which contains hundreds of videos of people expressing one of seven different emotions; Happy, Fear, Disgust, Surprise, Anger, and Contempt. Figure \ref{fig:emotions} illustrates the muscle movements that occur for these distinct emotions. We use the classification results of FAN \cite{meng2019frame} to weight our kernels.

In accordance with the facial action coding system that is used in \cite{ckplus}, our method clearly identifies the following action units in the Figure \ref{fig:emotions} videos which also correspond to their particular emotions: a lip corner puller in the happy video, a significant rise of the inner brow in the fear video, a nose wrinkler in the disgust video, an upper lip raiser and brow raisers in the surprise video, an inner brow raiser coupled with a lip corner depressor in the sad video, a tightening of the lips in the anger video, and a dimpling around the lips in the contempt video.
\begin{table}[!h]
  \centering
  \begin{tabular}{lc}
    \toprule
    Model & Average accuracy $\uparrow$ \\
    \midrule
    Face-GPS without FAN & 85.0\% \\
    Face-GPS with FAN & \textbf{86.1}\% \\
    \bottomrule
  \end{tabular}
  \caption{Classification results of facial muscle features on CK+ dataset.}
  \label{tab:results_comparison}
\end{table}

We present quantitative results to validate the effectiveness of our approach. We use an XGBoost classifier that is based solely on facial muscle displacements, without any visual information, to demonstrate its utility in classification tasks. Table~\ref{tab:results_comparison} reports the average classification accuracy on the CK+ dataset. Employing 10-fold cross-validation, the standalone Face-GPS method achieves an 85\% average accuracy on the test set, which increases to 86.1\% when enhanced with the FAN. This not only demonstrates the effectiveness of our kernel smoothing method but also shows that our facial muscle movements can serve as features for downstream tasks.

\section{Conclusions}

In this work, we present an end-to-end approach for dynamically quantifying facial muscle movements. Our method assesses these movements by tracking pixel displacements on a corresponding canonical face, allowing for accurate measurement even when the face is in motion or turned sideways. We develop a multi-kernel smoothing method to enhance the interpretability of face recognition deep learning models, highlighting the movements of specific muscle groups while filtering out noise from video recordings. Despite these advancements, capturing the facial manifold accurately, especially at its boundaries, remains a challenge and an area for future refinement. We also plan to improve this methodology to apply our kernels more precisely to the contours of facial muscles.

% \newpage
\section{Potential negative societal impact}
We foresee no negative societal impacts from enhancing our Face-GPS method, as it is intended to solely improve the explainability of deep-learning-based facial recognition models.

\section*{Acknowledgement}
We express our heartfelt thanks to Professor Miriam Rafailovich and her student Shi Fu for their guidance and expertise. Additionally, we are grateful to the participants of the Garcia Program, which allows gifted high school students to engage in independent research under the supervision of Garcia Center faculty and students.

{
  \small
  \bibliographystyle{unsrt}
  \bibliography{main}
}

%%%%%%%%%%%%%%%%%%%%%%%%%%%%%%%%%%%%%%%%%%%%%%%%%%%%%%%%%%%%

\end{document}